%% file: main.tex
\documentclass{article}

\usepackage{microtype}
\usepackage{graphicx}
\usepackage{xcolor}
\usepackage{booktabs}
\usepackage{longtable}
\usepackage{tabularx}
\usepackage{adjustbox}
\usepackage[utf8]{inputenc}
\usepackage[T1]{fontenc}
\usepackage{amsmath,amssymb,amsfonts}
\usepackage{hyperref}

\usepackage[preprint]{icml2026}
\setcitestyle{numbers,square,comma}

\icmltitlerunning{PaperJury: Due-Process Review for Bounded LaTeX Revision}

\begin{document}

\twocolumn[
\icmltitle{PaperJury: Due-Process Review for Bounded LaTeX Revision}

\icmlsetsymbol{corresponding}{\textdagger}

\begin{icmlauthorlist}
\icmlauthor{Yiran Wang*}{vast}
\icmlauthor{Ruixuan An*}{vast}
\icmlauthor{Biao Wu}{vast} 
\icmlauthor{Wenhao Wang}{vast,corresponding}
\end{icmlauthorlist}

\icmlaffiliation{vast}{Vast Intelligence Lab, Sydney, Australia}
\icmlcorrespondingauthor{Wenhao Wang}{wenhao.wang@example.com}
\icmlkeywords{Machine Learning, ICML}
\vskip 0.3in
]

\printAffiliationsAndNotice{}

\begin{abstract}
Pre-submission hardening of human-authored LaTeX computer science papers differs from drafting assistance because it requires adversarial whole-paper review, explicit no-fix outcomes, and bounded artifact-safe revision. Existing writing assistants, critique generators, and judge-centered loops lack durable issue identity across rounds, deterministic routing from critique to adjudication, and manuscript control that can reject invalid concerns or defer author-dependent ones. We present PaperJury, a closed-loop review-verdict-revise-verify system built on a deterministic-versus-semantic split: deterministic orchestration manages decomposition, a frozen claim spine, a durable ledger, routing, stopping, and exact-once patch application, while semantic agents are limited to bounded review, judgment, and repair. PaperJury combines bounded holistic review, contestability-based routing, a due-process trial, and risk-proportional guard chains for anchor-bounded edits, yielding terminal outcomes of invalid-drop, valid-fixable, and author-required. In a two-arm expert-review evaluation on held-out Vision, natural language processing, and machine learning papers against four baselines, we assess issue quality, verdict and routing quality, edit safety, convergence behavior, and cost, supporting the thesis that load-bearing safety and completion logic should reside in deterministic orchestration rather than model discretion. PaperJury is available at \url{https://github.com/u7079256/paperjury}.
\end{abstract}

\input{sections/introduction}

\input{sections/related_work}
\input{sections/method}

\input{sections/experiments}
\input{sections/conclusion}

\clearpage
\bibliographystyle{unsrtnat}
\bibliography{refs}


\end{document}

%% file: sections/introduction.tex
\section{Introduction}

A research paper can look finished long before its arguments are sound. Once the prose is polished, deeper problems become harder to notice rather than easier: a claim the experiments do not fully support, a contradiction between two sections, a number that suggests more than was actually measured~\cite{xu2026idea2story,qian2026story2proposal,xu2026selfevolving,dai2026papervoyager,dai2026iwebgenbench}. Consider an ordinary failure. An ablation table announces the best result on a benchmark, while three pages later a per-domain breakdown the authors built themselves quietly reports a higher number for a method they call a baseline. Each table is internally consistent, and the conflict surfaces only when one reader holds both in mind at once. Flaws like this are not typos a spell checker catches; they survive because they read fluently, and because each one is local while the evidence against it sits elsewhere in the manuscript. Finding them requires a critical reader who examines the whole manuscript and tries to challenge each important assertion, much as a careful referee would. Authors, however, rarely have access to such a reader before submission, and rereading their own draft makes it difficult to keep enough critical distance. As a result, strengthening a manuscript before submission remains a major bottleneck in scientific writing. Existing automated help was built for adjacent jobs: tools that refine wording or drive author-controlled style cleanup~\citep{ref_89dd03435fc867fa05945550d42232b4232e717b,ref_6e1db4a1906ba5e43ba1035929bab4029bf7f881}, and generators that produce focused critiques on local sections~\citep{ref_4636c68fe1fa65c311c38f72b9f7d2f72b4891dc}. These functions are useful, but none stands across the table, argues that a claim should not survive, and then carries that judgment through to a safe revision of the manuscript. Pre-submission hardening of a full \LaTeX{} manuscript is that missing job, and it is the problem we take up.

What makes the job hard is not writing the critique but governing what follows from it, under a precision-recall-cost trilemma: a useful system must find real problems, avoid inventing false ones, and do so within a bounded budget. A natural shortcut is to use large language models (LLMs) as automatic judges, but this fails exactly here, because judgments from a single model can change when the prompt is rewritten~\citep{ref_16f1651153c5661719aa487fc73c8e9119cf27cc} and automated reviews still suffer from repeated, aspect-specific failures~\citep{ref_455ec77b8eaf8ee4cd54a8f6bd9226b0e75df87e,ref_4ba68fdaab6ee7b6b32088e96876d329646631a0,ref_995bba3d9c845b1cc88f9c2baa60a7c2b283bc78}. Even richer systems that deepen a single automated reviewer~\citep{ref_60c8a127e6ae8c8e21dd7edfc187ff7f0d9ae2bd}, stage courtroom-style debate among assigned roles~\citep{ref_0ff27b835c06ab61b743d8596673720558a08ada}, or validate each model step against typed constraints~\citep{d61b48c5b4086649b9bd0ff5b61f9c626bcf8f2b} still leave the safety-critical decisions to the model, and those decisions fall into two challenges. The first is \emph{adjudication}: which complaints to honor, and when to stop. Many machine-generated complaints are incorrect, such as a hallucinated contradiction between two passages that in fact agree, so a loop that acts on everything it hears corrupts a sound paper; many correct complaints instead require new experiments that software cannot perform, so a binary fix-or-ignore decision is the wrong shape for the problem; and a model asked whether it is finished tends to say yes, because the component that proposes changes is the worst possible certifier of its own completion. The second is \emph{revision}: where an edit may safely land. A patch meant to tighten one sentence can silently change a neighboring claim, break a cross-reference, or contradict the abstract two sections away, and a model editing free-hand has no reliable way to bound its own blast radius, so throughput bought by automated rewriting is paid for in unsafe edits~\citep{bc4945c17eae91b1333b531296178f6a7847aa05}.

\begin{figure*}[t]
  \centering
  \includegraphics[width=1\textwidth]{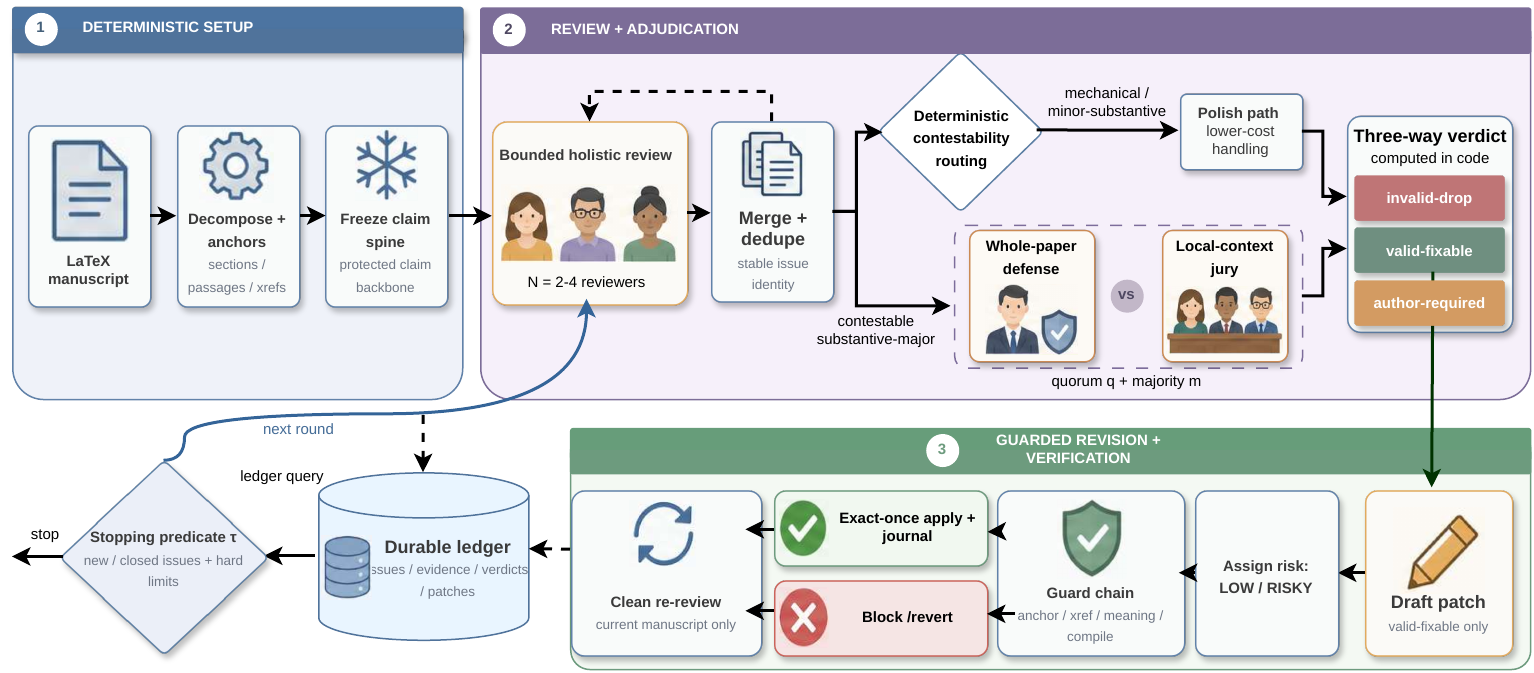}
  \caption{Overview of the PaperJury review-verdict-revise-verify pipeline.}
  \label{fig:paperjury_pipeline}
  \vspace{0.5cm}
\end{figure*}

To address this, we present PaperJury, a closed-loop system for reviewing, judging, revising, and verifying \LaTeX{} computer-science manuscripts. Its central design principle is a strict separation between deterministic control and semantic reasoning: all operations that must behave consistently across runs are handled by deterministic code, while language models are used only for bounded tasks that require reading, reasoning, judgment, or drafting. This separation answers the two challenges in turn (Figure~\ref{fig:paperjury_pipeline}). Against unreliable \emph{adjudication}, the decisions of which complaints to honor and when to stop move out of the model and into deterministic orchestration. A durable ledger gives each issue a stable identity across rounds, so a complaint persists as the same tracked object while the models around it are re-run. Each issue is routed by a deterministic contestability rule rather than a model's mood, sending costly adjudication only where a mistaken first call would hurt most; contestable issues enter a due-process trial in which a whole-paper defense argues against the charge while a decorrelated local-context jury evaluates it using localized evidence, so the two views cannot contaminate each other. Because issue validity does not imply machine editability, every issue terminates in one of three outcomes, invalid-drop, valid-fixable, or author-required, and the loop stops on a deterministic ledger query rather than a model's sense that it is done. Against unsafe \emph{revision}, the system first decomposes the manuscript and extracts a frozen claim spine that records its central assertions and prevents them from drifting, so an edit may repair support or wording but never silently rewrite a claim. Only valid-fixable issues are edited, and each fix is an anchor-bounded edit screened by a risk-proportional guard chain that verifies the edit stays within its intended scope and does not silently change neighboring claims, references, or results, before it is applied exactly once and journaled so it can be reverted. The model still does the hard reading; it no longer holds the gavel or the eraser.

We make three contributions, aligned with the two challenges and their evaluation:
\begin{itemize}
\item{Governed adjudication.} We cast pre-submission hardening as a closed adversarial loop and move its load-bearing calls, which complaints to honor and when to stop, out of the model into deterministic orchestration whose verdicts can drop an issue, fix it, or defer it to the author.

\item{Artifact-safe revision.} We make automated editing safe by construction, so every applied fix is confined to its claimed span and remains reversible.

\item {Evaluation and release.} We evaluate PaperJury against four baselines in a two-arm expert study over twelve vision, natural language processing, and machine learning papers, and release the full implementation.
\end{itemize}


%% file: sections/related_work.tex
\section{Related Work}

\subsection{Review and Revision Systems  }

AI support for scientific writing has largely emphasized feedback generation and draft improvement rather than pre-submission hardening over a full LaTeX manuscript. XtraGPT treats academic writing as iterative revision and emphasizes context-aware, criteria-guided editing for scientific drafts, making it a close revision-side antecedent, while also stressing cross-section coherence during revision; however, it does not provide artifact-safe LaTeX revision with a frozen claim spine, anchor-bounded edits, or orchestrator-controlled apply and revert semantics~\cite{ref_89dd03435fc867fa05945550d42232b4232e717b}. SWIF$^{2}$T generates focused, actionable feedback on paper weaknesses, but remains a feedback generator rather than a system that decides whether an issue should be dropped, fixed, or deferred to the author~\cite{ref_4636c68fe1fa65c311c38f72b9f7d2f72b4891dc}. TreeReview strengthens full-paper review through hierarchical question decomposition and dynamic follow-up expansion, improving the depth and efficiency of review generation~\cite{fccde18f25977d129af241996a769bdb23d1148d}. Related rebuttal-oriented systems such as DEFEND also keep the author in the loop for targeted response generation, reinforcing that prior tools assist with downstream review interaction rather than providing a governed pre-submission hardening loop over the manuscript itself~\cite{f82d8c1877b1e775e514cdb1ba539d7689c8f4ee}. Taken together, these systems offer stronger critique, revision, review, or rebuttal primitives, but they still operate in a forward critique-or-rewrite mode rather than a closed-loop process with durable issue identity, cross-round state, no-fix-capable outcomes, and deterministic edit-safety controls; thus, prior review and revision systems do not combine bounded holistic review, adjudication into invalid-drop, valid-fixable, or author-required outcomes, and deterministic edit-safety controls within one governed loop.

\subsection{Guardrailed Agent Workflows}

A second line of work contributes ingredients for deliberation, judging, and workflow control, but does not place safety-critical authority in deterministic orchestration for manuscript revision. Studies of LLM-as-a-Judge report systematic scoring and position biases, cautioning against treating a single semantic verdict as a reliable final decision in load-bearing workflows~\cite{e142887ab6516654f25e233a3e661eddac123630,d0b26fc297c2779b4c59ae019dc8750f614c3f47,dfbfe75ec8c2143e899897a3c054ee58d99ead43}. PROClaim introduces courtroom-style adversarial roles, including defense and judge, and uses aggregation to support controversial decisions~\cite{ref_0ff27b835c06ab61b743d8596673720558a08ada}. TraceSafe shows that the main safety surface in agentic workflows lies in intermediate execution traces and that effective guarding depends heavily on structural validation, while ATLAS likewise embeds LLM generation in a layered workflow separating representation, constraint compilation, and post-generation validation so failures become explicit and diagnosable~\cite{bc4945c17eae91b1333b531296178f6a7847aa05,d61b48c5b4086649b9bd0ff5b61f9c626bcf8f2b}. TreeReview offers an efficiency-aware structure for review generation, and multi-agent peer-review frameworks such as ScholarPeer similarly show how agent specialization can improve critique production, but efficiency and specialization alone do not determine whether contested issues should be overruled, revised, or escalated to the author~\cite{fccde18f25977d129af241996a769bdb23d1148d,ref_60eabd4ff812051197450a5b1843f1e3fea723db}. These works motivate adversarial adjudication and validation-backed automation, yet they do not provide deterministic routing from critique to due-process trial, a durable ledger across rounds, or risk-proportional revision tied to a frozen claim spine. PaperJury sits between these two lines: it addresses scientific-paper review and revision like XtraGPT~\cite{ref_89dd03435fc867fa05945550d42232b4232e717b} while adopting the control and validation concerns emphasized by work on judging and agent workflows. The remaining gap is a closed-loop pre-submission review architecture that combines bounded whole-paper issue generation, deterministic orchestration, explicit terminal outcomes, and artifact-safe LaTeX revision in one system.

%% file: sections/method.tex
\section{Method}

\begin{figure*}[t]
  \centering
  \includegraphics[width=0.9\textwidth]{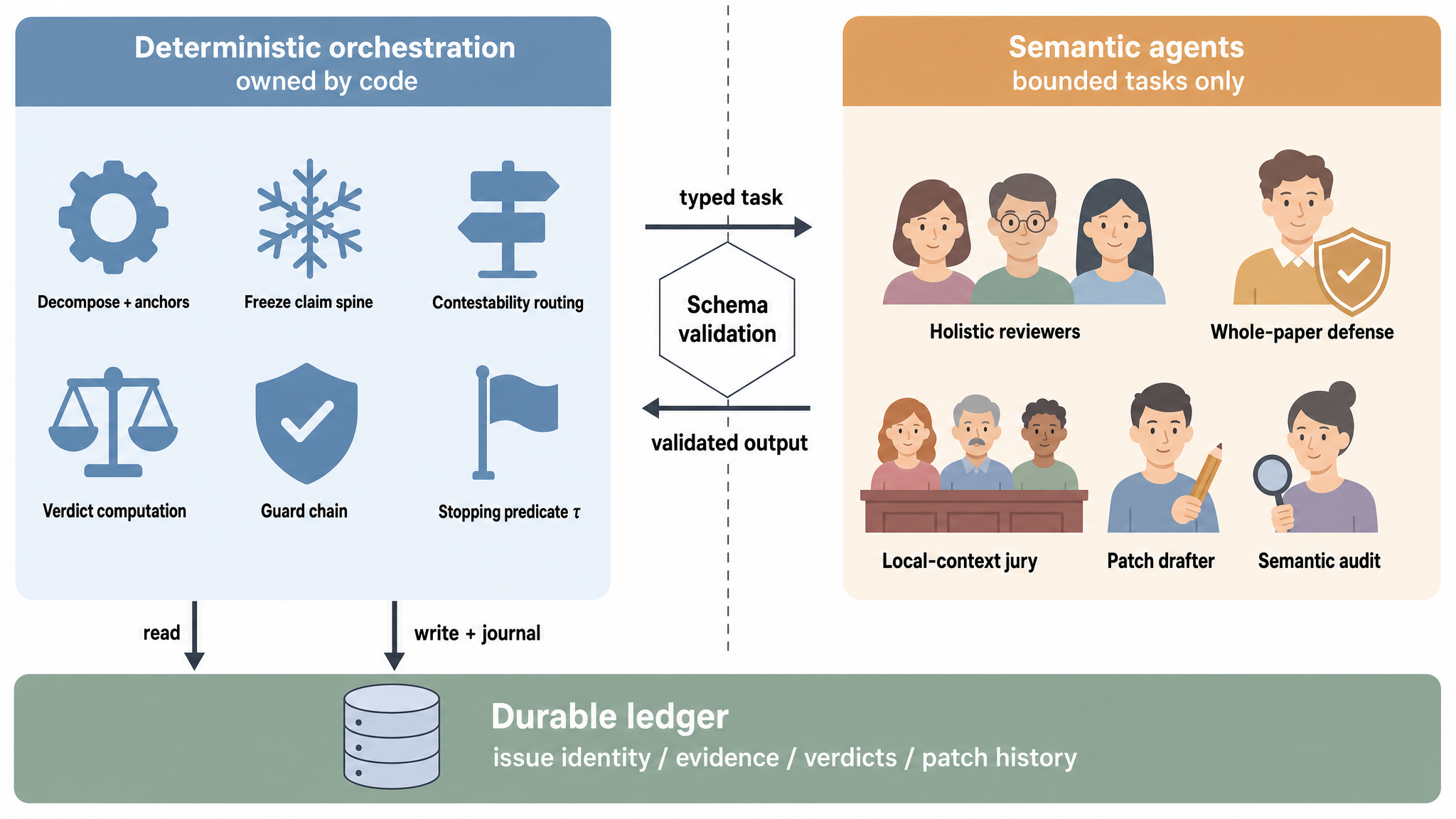}
  \caption{Deterministic-versus-semantic split with ledger-backed control flow.}
  \label{fig:deterministic_semantic_split}
\end{figure*}

PaperJury is a closed-loop pre-submission review system for LaTeX computer science papers. Its central design principle is a deterministic-versus-semantic split: deterministic orchestration owns state transitions, guards, stopping, and exact-once patch application, while semantic agents are limited to bounded review, judgment, drafting, and audit tasks. This places model generation inside an explicitly controlled workflow rather than letting models execute the protocol itself \cite{d61b48c5b4086649b9bd0ff5b61f9c626bcf8f2b}, a choice motivated by evidence that the dominant safety surface of agentic execution lies in the models' intermediate execution traces \cite{bc4945c17eae91b1333b531296178f6a7847aa05}. Figure \ref{fig:paperjury_pipeline} summarizes the review-verdict-revise-verify pipeline, and Figure \ref{fig:deterministic_semantic_split} shows the separation between deterministic orchestration and semantic agents.

\subsection{Problem Formulation}

Given an input LaTeX manuscript $x$, PaperJury returns a durable ledger of candidate issues $I$, terminal verdicts $\{v_i\}_{i \in I}$, and, when permitted, an exact-once applied edit set $A$. The system is designed around a precision-recall-cost trilemma: broader issue generation may improve recall but increases token and adjudication cost, while aggressive automated revision may increase throughput but also increase the risk of unsafe edits. PaperJury addresses this by bounding whole-paper issue generation, escalating only contestable substantive-major issues to expensive adjudication, and allowing explicit no-fix-capable outcomes when machine revision is unwarranted.

The outer loop alternates among review, ledger merge, adjudication, guarded revision, and clean re-review. Candidate issues may terminate as invalid-drop, valid-fixable, or author-required, because issue validity does not imply machine editability and some credible concerns should preserve author intent rather than trigger rewriting. Stopping is determined by a deterministic ledger-query predicate rather than model self-assessment, so the same semantic component does not both propose changes and certify completion~\cite{dai2026iwebgenbench,shi2025presentagent,dai2026papervoyager}.

\subsection{Notation}

We define the key symbols used throughout the method in Table~\ref{tab:notation}.

\begin{table*}[t]
\centering
\small
\caption{Notation used throughout the method and evaluation.}
\label{tab:notation}
\renewcommand{\arraystretch}{0.92}
\setlength{\tabcolsep}{4pt}
\begin{tabularx}{\textwidth}{@{}l X @{\hspace{1.5em}} l X@{}}
\toprule
Symbol & Meaning & Symbol & Meaning \\
\midrule
$x$ & input LaTeX manuscript & $G(P_i)$ & guard-chain outcome for patch $P_i$ (anchor, reference, semantic, compile checks) \\
$D(x)$ & deterministic decomposition of $x$ into sections, passages, anchors, and cross-reference targets & $A$ & set of applied valid-fixable edits after exact-once patching \\
$S$ & frozen claim spine extracted from the manuscript & $U_{r}$ & genuinely new issues discovered in round $r$ \\
$K$ & outer-loop review rounds until termination (five-round cap) & $C_{r}$ & issues closed in round $r$ \\
$N$ & number of holistic domain reviewers, clamped to $[2,4]$, default 3 & $\tau$ & deterministic stopping predicate for unattended execution \\
$I$ & set of candidate issues produced for a manuscript & $M_{h}$ & expert human issue panel used as reference standard \\
$i$ & a single candidate issue & $M_{p}$ & PaperJury issue set and verdict outputs \\
$L$ & durable ledger storing issue state, evidence, verdicts, and application history & $P_{\mathrm{panel}}$ & panel-relative precision of major issues under fix-equivalence matching \\
$e_{i}$ & evidence anchors and supporting passages associated with issue $i$ & $P_{\mathrm{verified}}$ & audit-corrected precision crediting expert-validated system-only issues \\
$c_{i}$ & contestability label for issue $i$ used by deterministic routing & $R$ & panel-relative recall of major issues \\
$v_{i}$ & terminal verdict for issue $i$ in \{invalid-drop, valid-fixable, author-required\} & $F1$ & per-paper harmonic mean of precision and recall, macro-averaged \\
$T(i)$ & trial procedure invoked for routed issue $i$ & $\mathrm{Acc_v}$ & verdict accuracy against expert labels \\
$J$ & jury panel used in two-sided adjudication & $\mathrm{Acc_r}$ & routing accuracy against expert-endorsed decisions \\
$q$ & quorum threshold for deterministic verdict computation & $\mathrm{ESVR}$ & edit-safety violation rate over applied edits \\
$m$ & majority threshold for deterministic verdict computation & $W$ & wall-clock runtime per paper, in hours \\
$P_{i}$ & proposed patch for issue $i$ & $B_{\mathrm{naive}}$ & naive unbounded per-(unit×lens) review baseline \\
$a(P_i)$ & anchor-bounded diff induced by patch $P_i$ & $C_{\mathrm{attn}}$ & compute cost in agents spawned, tokens, and runtime \\
$\rho_{i}$ & risk category of patch $P_i$, e.g., LOW or RISKY & $r$ & outer-loop round index, $r \in \{1, \dots, K\}$ \\
\bottomrule
\end{tabularx}
\end{table*}

Let $D(x)$ denote deterministic decomposition of manuscript $x$ into sections, passages, anchors, and cross-reference targets, and let $S$ denote the frozen claim spine extracted before revision. Let $N$ be the number of holistic domain reviewers, clamped to $[2,4]$ and defaulted to $3$. Review produces candidate issues $I$, where each issue $i \in I$ has evidence anchors $e_i$ and a contestability label $c_i$. A durable ledger $L$ stores issue identity, evidence, verdicts, and patch history. Routed issues may enter a due-process trial $T(i)$ with jury panel $J$, quorum threshold $q$, and majority threshold $m$, producing $v_i \in \{\text{invalid-drop}, \text{valid-fixable}, \text{author-required}\}$.

For revision, let $P_i$ denote a proposed patch for issue $i$, $a(P_i)$ its anchor-bounded diff, and $G(P_i)$ the guard-chain outcome. Let $\rho_i$ denote the patch risk category, with discrete labels such as LOW and RISKY. The exact-once applied edit set is $A$. Across rounds $r \in \{1, \dots, K\}$, let $U_r$ be newly discovered issues and $C_r$ be closed issues. The unattended loop stops when a deterministic predicate $\tau$ over ledger queries is satisfied.

\subsection{Review and Adjudication Pipeline}

The pipeline begins with deterministic decomposition through $D(x)$ and extraction of the frozen claim spine $S$. This stage creates stable addressable units before any semantic agent is invoked, tying later review, adjudication, and revision to persistent anchors rather than ad hoc text spans. The frozen claim spine is the protected semantic backbone for later guard decisions: machine edits may repair support, wording, or local organization, but they must not silently alter claim-level commitments. This follows the broader design of placing generation inside a typed, validated workflow rather than leaving structure and state identity to unconstrained model behavior \cite{d61b48c5b4086649b9bd0ff5b61f9c626bcf8f2b}.

PaperJury then runs bounded holistic review with a small set of whole-paper reviewers. The reviewer count $N$ is clamped to $[2,4]$ and defaults to $3$. Each reviewer reads the full manuscript once and emits evidence-anchored weaknesses with coverage signals, since unsupported positioning, cross-section coherence failures, and missing experimental justification often emerge only under whole-paper inspection. PaperJury bounds this stage to control the precision-recall-cost trilemma: unlike an unbounded critique generator that fans out by unit and lens, it limits first-pass issue generation and invokes targeted re-read only when deterministic anti-skim checks indicate missed coverage or weak grounding, echoing the need-based deeper probing of efficiency-aware review generation \cite{fccde18f25977d129af241996a769bdb23d1148d}.

Reviewer outputs are merged into the durable ledger, where duplicate or near-duplicate weaknesses are consolidated and evidence provenance is retained. Because issues are grounded in deterministic decomposition, durable issue identity remains stable across rounds and supports later routing, closure, and exact-once revision.

\begin{figure*}[t]
  \centering
  \includegraphics[width=0.9\textwidth]{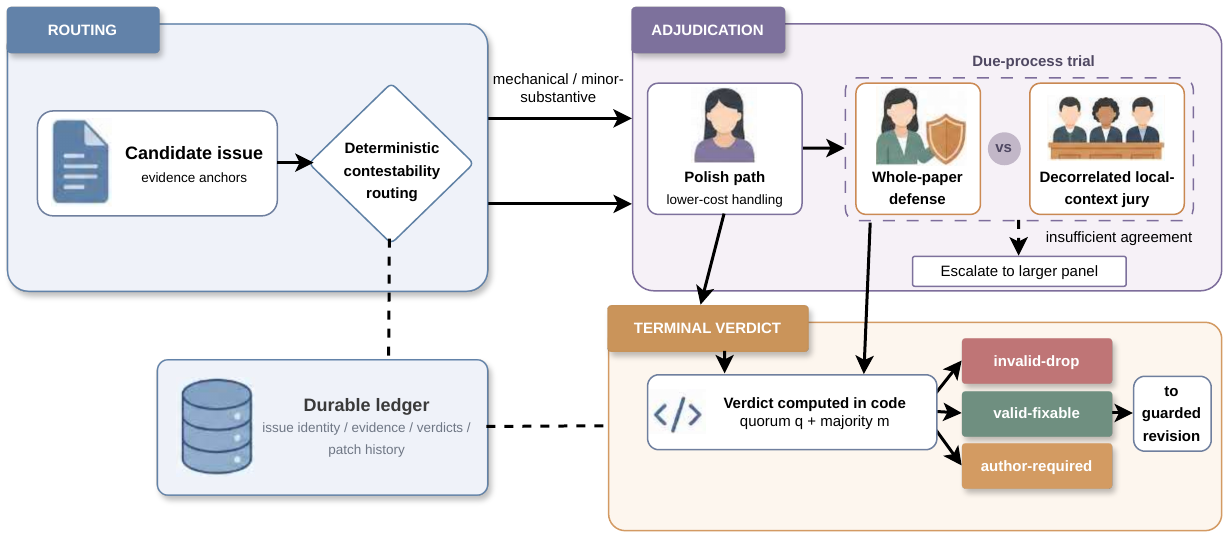}
  \caption{Deterministic routing and due-process adjudication for candidate issues.}
  \label{fig:routing_and_trial}
\end{figure*}

After merge, PaperJury assigns each issue a deterministic contestability label $c_i$ and routes it accordingly. Mechanical and minor-substantive issues follow a lower-cost polish path, while contestable substantive-major issues enter a due-process trial, illustrated in Figure \ref{fig:routing_and_trial}. Routing is deterministic because adjudication cost should be concentrated on the cases where mistaken first-pass judgment is most consequential.

For a routed issue $i$, the due-process trial combines whole-paper defense with a decorrelated local-context jury. The defense argues against the charge using manuscript-wide context, while the jury evaluates localized evidence with reduced dependence on the original reviewer framing. If agreement is insufficient, the issue may escalate to a larger panel. Final verdicts are computed in code using quorum and majority rules:
\begin{equation}
v_i = \mathrm{Verdict}(T(i), q, m).
\end{equation}

This design rejects giving sole authority to a single semantic judge: prior work shows that LLM-as-a-Judge behavior can shift under prompt perturbations and related presentation effects, and related multi-agent arbitration results likewise support replacing a single semantic judge with structured debate and group decision procedures for ambiguous contested cases~\cite{ref_16f1651153c5661719aa487fc73c8e9119cf27cc,ref_4996e65fad20d96ae919d5aba2be8060d4a39003}. The resulting verdict space is three-way by construction: invalid-drop for rejected charges, valid-fixable for upheld issues that are safely machine-editable, and author-required for upheld issues that should not be automatically rewritten. PaperJury therefore separates issue validity from editability and preserves no-fix-capable outcomes.

\subsection{Guarded Revision and Convergence}

Only issues with $v_i = \text{valid-fixable}$ proceed to revision. For each such issue, a drafter proposes a patch $P_i$, which is screened by a risk-proportional guard chain before any text is committed. The guard chain includes anchor-bounded diff checks, cross-reference checks, semantic meaning or edit audits when triggered, compile checks, and application journaling. Let
\begin{equation}
G(P_i) \in \{\text{pass}, \text{fail}\}
\end{equation}
denote the aggregate outcome.

Risk-proportional revision uses deterministic prefilters on $a(P_i)$ and related structural signals to assign a risk category $\rho_i$, such as LOW or RISKY. LOW patches follow a lighter semantic path, while RISKY patches trigger stronger audits. This avoids sending every patch through the same expensive verifier while also refusing to trust superficially plausible prose when structural failures such as reference breakage, anchor drift, or silent claim mutation remain possible. The layered design follows the same bounded-automation pattern used in structured artifact workflows~\cite{d61b48c5b4086649b9bd0ff5b61f9c626bcf8f2b,c2217289dd5624e66a665f9d2e549c1a0c259c9d}.

The frozen claim spine $S$ is the key revision constraint. If a patch appears to move beyond local repair toward claim-level alteration, it is surfaced as risky and may be blocked or escalated. Cross-reference and compile guards further preserve document integrity by preventing broken references and non-compiling LaTeX. If a patch passes all guards, it is applied exactly once and journaled in the ledger:
\begin{equation}
A = \{P_i \mid v_i = \text{valid-fixable} \land G(P_i)=\text{pass}\}.
\end{equation}
If later checks fail, the orchestrator can revert through the journaled state, avoiding duplicate edits, inconsistent local states, and divergence between manuscript text and ledger history \cite{ref_28f570c7445be4eaaa18f1c6eb109ec8e37d9dea}.

The procedure can be summarized compactly. The orchestrator decomposes the manuscript, freezes the claim spine, initializes the durable ledger, runs bounded holistic review with anti-skim re-read if needed, merges issues, applies deterministic routing, computes verdicts through polish or due-process trial, drafts patches only for valid-fixable issues, assigns patch risk, executes the guard chain, applies or reverts through exact-once patching, then performs clean re-review and ledger updates until the stopping predicate is met.

After guarded edits, PaperJury performs clean re-review on the current manuscript state rather than asking agents to continue from their earlier outputs. This is a design motivation for reducing carryover from prior critiques and for surfacing genuinely new issues introduced by edits. The outputs are merged by a deterministic clerk into the cumulative ledger.

Termination is defined by a deterministic ledger-query predicate over newly discovered and closed issues. Let $U_r$ denote genuinely new issues in round $r$ and $C_r$ the issues closed in that round. The unattended loop halts when
\begin{equation}
\tau(L, U_r, C_r, r) = \text{true},
\end{equation}
where $\tau$ is a stopping rule over ledger state and hard execution limits. Convergence is therefore certified by explicit state queries rather than by a model declaring the manuscript complete. Together, bounded holistic review, due-process adjudication, risk-proportional guard chains, exact-once patch application, and deterministic outer-loop convergence yield artifact-safe LaTeX revision under deterministic orchestration.

%% file: sections/experiments.tex
\section{Experiments}

\subsection{Implementation Details}

PaperJury is implemented as deterministic orchestration over schema-validated semantic agents for full LaTeX computer science manuscripts. Decomposition, durable ledger updates, contestability-based routing, the ledger-query predicate, exact-once patch application, journaling, and revert semantics run in deterministic code, while semantic agents are limited to bounded review, judgment, drafting, and audit subtasks. Unless noted otherwise, bounded holistic review uses three reviewers, with reviewer count clamped to two to four, and the outer loop proceeds through clean re-review rounds until termination, subject to a five-round execution cap. All methods are evaluated on the same held-out paper set with the same manuscript artifact, review-budget accounting interface, and logging of spawned agents, token use, and wall-clock time. For revision-capable systems, preprocessing preserves LaTeX structure and records anchors for anchor-bounded edits; guarded variants use compile-aware validation under orchestrator-controlled apply and revert semantics. Each round re-reads the current manuscript rather than prior model outputs, and baselines are run under matched prompt framing and stopping-budget envelopes when their native design does not define deterministic completion.

\subsection{Experimental Design}

We use a two-arm expert-review evaluation on a held-out corpus of LaTeX computer science papers spanning vision, natural language processing, and machine learning. Arm 1 compares system-discovered issues with expert issue panels using passage-level and semantic matching, while verdict quality is audited separately by terminal class (invalid-drop, valid-fixable, author-required) so that issue quality is not conflated with downstream adjudication quality, which is itself known to be hard to estimate reliably with LLM judges~\cite{ref_9c9eb9d73333ba9ecbfcbe7cf1e0a8971f9222c2,ref_35face12ebc1d904e67d2ee4e0be53b7c84ffc96}. Arm 2 asks the same experts to audit verdicts, routed cases, applied edits, and dropped issues, yielding measures of verdict correctness, routing quality, edit safety, and whether stopping behavior matches expert judgment. We compare PaperJury with four baselines: Forward-Only Rewriter; LLM Critic Only; LLM-as-Judge Review-Revise Loop, which relies on model judgment for adjudication and stopping despite known reliability concerns in judge-centered pipelines~\cite{ref_9c9eb9d73333ba9ecbfcbe7cf1e0a8971f9222c2,ref_35face12ebc1d904e67d2ee4e0be53b7c84ffc96,febde88848d1d1f95da08737552c1056a9e79565,ref_546e402317e1291613a8848e1efbc683f2968d7a}; and Naive Unbounded Per-(Unit$\times$Lens) Generator, a high-recall, high-cost reference for exhaustive critique expansion, in contrast to structured review pipelines \cite{ref_60c8a127e6ae8c8e21dd7edfc187ff7f0d9ae2bd}. We report issue quality, verdict correctness, routing quality, edit safety, convergence behavior, and cost in tokens and wall-clock time, since iterative multi-agent systems should be evaluated jointly on quality, stopping, and efficiency \cite{d78ea99ff3ab095a6ee03deb3a6a7c95501124ed}. \begin{figure}[t]
  \centering
  \includegraphics[width=0.92\columnwidth]{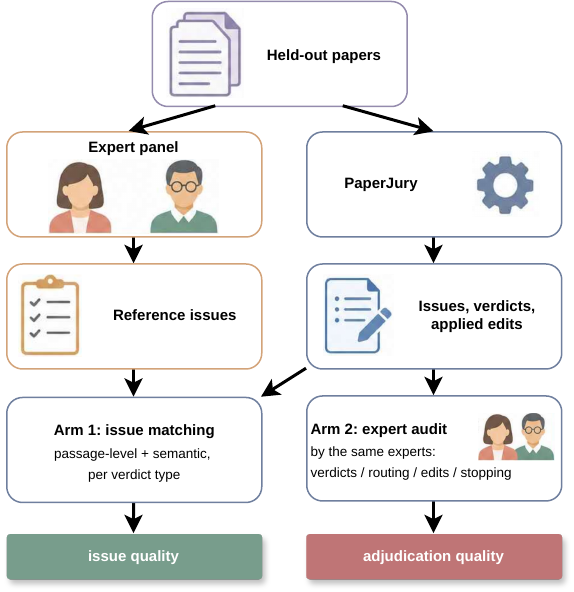}
  \caption{Two-arm expert-review evaluation protocol for PaperJury.}
  \label{fig:eval_protocol}
  \vspace{-0.08cm}
\end{figure}
\begin{figure}[t]
  \centering
  \includegraphics[width=0.92\columnwidth]{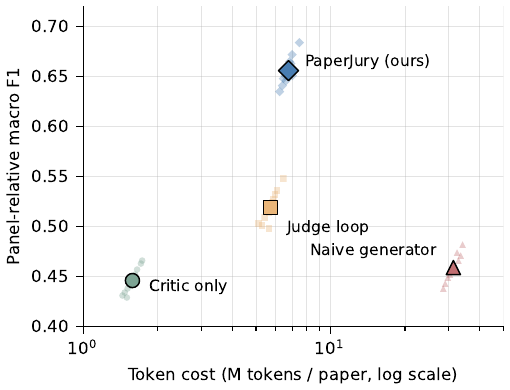}
  \caption{Quality-cost trade-off across issue-producing systems: per-paper faint dots and system means, with cost measured in millions of tokens per paper (log scale). The forward-only rewriter produces no issue list, hence no F1, and is excluded from this frontier.}
  \label{fig:cost_quality_tradeoff}
\end{figure}
Figure~\ref{fig:eval_protocol} summarizes the separation between issue discovery and downstream auditing, and Figure~\ref{fig:cost_quality_tradeoff} visualizes the cost-quality tradeoff relative to the exhaustive generator.

\subsection{Results}

Throughout, P\textsubscript{panel} and R denote panel-relative precision and recall of system-discovered major issues matched against the expert issue panels under fix-equivalence; F1 is computed per paper as the harmonic mean of that paper's precision and recall, then macro-averaged across papers. P\textsubscript{verified} additionally credits system-only issues that blinded experts judge valid in the audit arm; it is reported with its audited sample size and is never used inside F1. Acc~v and Acc~r denote blinded expert agreement with the system's terminal verdicts and routing decisions; audit items that experts mark cannot-tell are excluded from the agreement denominator and reported separately. ESVR is the number of safety-violating applied edits over all applied edits, pooled across papers, and is always read together with edit volume and coverage (Table~\ref{tab:esvr_companion}). K is the number of review rounds to termination, reported as mean\,$\pm$\,sd across papers together with the fraction of papers hitting the five-round execution cap, and W is wall-clock time per paper in hours. All 95\% confidence intervals are Wilson score intervals on audited counts. Cells marked n/a correspond to capabilities a method does not have (e.g., no explicit issue list, no verdicts, or no applied edits).

\begin{table*}[t]
\centering
\small
\caption{Main end-to-end results across PaperJury and four baselines on issue quality, adjudication, edit safety, and efficiency. Metric and convention definitions are given at the start of this section; ESVR is read with Table~\ref{tab:esvr_companion}.}
\label{tab:main_results}
\setlength{\tabcolsep}{3.5pt}
\begin{tabular}{lcccccccccc}
\toprule
Method & P\textsubscript{panel}/P\textsubscript{verified} & n & 95\% CI & R & F1 & Acc v & Acc r & ESVR & K & W \\
\midrule
Forward-only rewriter & n/a & n/a & n/a & n/a & n/a & n/a & n/a & 0.240 & 1 & 0.31 \\
LLM critic only & 0.437\,/\,0.577 & 78 & [0.466, 0.680] & 0.462 & 0.446 & n/a & n/a & n/a & 1 & 0.51 \\
LLM-as-judge loop & 0.512\,/\,0.663 & 92 & [0.562, 0.751] & 0.533 & 0.519 & 0.681 & n/a & 0.110 & 3.33\,$\pm$\,1.07 (2/12) & 2.06 \\
Naive unbounded generator & 0.341\,/\,0.511 & 92 & [0.410, 0.611] & 0.721 & 0.459 & n/a & n/a & n/a & 1 & 8.37 \\
PaperJury (ours) & 0.684\,/\,0.847 & 98 & [0.763, 0.905] & 0.637 & 0.656 & 0.887 & 0.913 & 0.025 & 3.08\,$\pm$\,0.67 (0/12) & 2.47 \\
\bottomrule
\end{tabular}
\end{table*}
Table~\ref{tab:main_results} gives the main end-to-end comparison across PaperJury and the four baselines on issue matching, verdict correctness, routing correctness, edit-safety violations, convergence behavior, and cost. The key comparison is not any single metric in isolation, but whether a system can sustain issue quality while preserving governed adjudication, artifact-safe revision, and practical efficiency. Read with Figure~\ref{fig:cost_quality_tradeoff}, the table positions PaperJury against two failure modes: cheaper but less governed pipelines such as Forward-Only Rewriter and LLM Critic Only, and higher-cost expansion from Naive Unbounded Per-(Unit$\times$Lens) Generator. LLM-as-Judge Review-Revise Loop is the closest judge-centered comparator because it couples critique, judgment, and stopping inside semantic decisions rather than deterministic control. PaperJury attains the best panel-relative F1 (0.656, vs.\ 0.519 for the judge-centered loop), the highest audited precision (0.847 on 98 audited issues), verdict and routing agreement of 0.887 and 0.913, and the lowest ESVR among edit-applying systems (0.025), while terminating in about three rounds (3.08\,$\pm$\,0.67, never hitting the cap) at 2.47 hours and 6.76 million tokens per paper. The naive generator reaches the highest recall (0.721) but with the lowest precision (0.341) at 8.37 hours and 31.4 million tokens per paper, and the judge-centered loop hits the round cap on 2 of 12 papers, consistent with less stable self-stopping. Per-paper paired F1 differences favor PaperJury on all 12 papers against every issue-producing baseline; paper-cluster paired-bootstrap 95\% lower bounds on $\Delta$F1 are $+0.200$ against the critic, $+0.127$ against the judge-centered loop, and $+0.188$ against the naive generator, satisfying the preregistered primary endpoint.

\begin{table}[t]
\centering
\small
\caption{Verdict audit by terminal class: blinded experts relabel sampled ledger issues without seeing the system's verdict; cannot-tell items are excluded from the denominator and reported separately.}
\label{tab:verdict_audit}
\setlength{\tabcolsep}{3pt}
\resizebox{\columnwidth}{!}{%
\begin{tabular}{lcccccc}
\toprule
Terminal class & n & Papers & Acc v & 95\% CI & cannot-tell & Main confusion \\
\midrule
invalid-drop & 43 & 12 & 0.872 & [0.733, 0.944] & 4 (9.3\%) & inv$\to$VF \\
valid-fixable & 73 & 12 & 0.913 & [0.823, 0.960] & 4 (5.5\%) & VF$\to$AR \\
author-required & 51 & 11 & 0.860 & [0.727, 0.934] & 8 (15.7\%) & AR$\to$VF \\
\bottomrule
\end{tabular}%
}
\end{table}

\begin{table}[t]
\centering
\small
\caption{PaperJury per-domain slices (four papers per family; diagnostic, not powered comparisons). The three audited slices ($n=31+35+32$) compose the overall audit in Table~\ref{tab:main_results}.}
\label{tab:domain_slices}
\setlength{\tabcolsep}{5pt}
\resizebox{\columnwidth}{!}{%
\begin{tabular}{lcccccccc}
\toprule
Domain & P\textsubscript{panel}/P\textsubscript{verified} & n & 95\% CI & R & F1 & Acc v & Acc r & ESVR \\
\midrule
Vision & 0.671\,/\,0.839 & 31 & [0.674, 0.929] & 0.626 & 0.646 & 0.881 & 0.904 & 0.028 \\
NLP & 0.701\,/\,0.857 & 35 & [0.706, 0.937] & 0.653 & 0.671 & 0.902 & 0.929 & 0.021 \\
ML & 0.680\,/\,0.844 & 32 & [0.682, 0.931] & 0.632 & 0.651 & 0.878 & 0.906 & 0.026 \\
\bottomrule
\end{tabular}%
}
\end{table}

Table~\ref{tab:verdict_audit} audits the three-way terminal verdict space directly. Verdict slices cannot carry panel-matching precision or recall: a correctly dropped invalid issue should not match the expert panel at all, and unmatched panel issues hold no terminal class, so the appropriate per-class evidence is blinded relabeling agreement together with its confusion structure. Performance on valid-fixable cases alone would not justify a three-way terminal verdict space, so the invalid-drop and author-required rows carry the burden; both hold up (0.872 and 0.860, against 0.913 for valid-fixable), and the measured errors concentrate on the valid-fixable--author-required boundary, where the boundary class author-required also has the highest cannot-tell rate (15.7\%), reflecting defensible disagreement rather than adjudication failure. Table~\ref{tab:domain_slices} then tests whether discovery and adjudication quality hold across Vision, natural language processing, and machine learning papers rather than being concentrated in one subcommunity; with four papers per family, these slices are diagnostic, and all metrics stay within 0.03 of the pooled values. Convergence behavior appears in the K column of Table~\ref{tab:main_results}: PaperJury terminates deterministically in 3.08\,$\pm$\,0.67 clean re-review rounds without hitting the five-round cap, whereas the judge-centered loop self-stops at 3.33\,$\pm$\,1.07 rounds and hits the cap on 2 of 12 papers; analyzing round-level convergence rather than single-pass outputs follows stability-based stopping analyses for iterative multi-agent judging \cite{d78ea99ff3ab095a6ee03deb3a6a7c95501124ed}.

\begin{table*}[t]
\centering
\small
\caption{Edit-safety companion to Table~\ref{tab:main_results}: applied-edit volume, coverage of valid-fixable terminals, guard-block rate, and unsafe edits per system. The critic-only and naive baselines apply no edits.}
\label{tab:esvr_companion}
\setlength{\tabcolsep}{4pt}
\begin{tabular}{lcccccccc}
\toprule
Method & Applied (per paper) & Papers w/ edits & VF terminals & Proposed & Coverage & Guard-block & Unsafe & ESVR \\
\midrule
Forward-only rewriter & 254 (21.2) & 12/12 & n/a & n/a & n/a & n/a & 61 & 0.240 \\
LLM-as-judge loop & 172 (14.3) & 12/12 & 204 & 178 & 0.843 & 0.034 & 19 & 0.110 \\
PaperJury (ours) & 161 (13.4) & 12/12 & 196 & 194 & 0.821 & 0.170 & 4 & 0.025 \\
\bottomrule
\end{tabular}
\vspace{3mm}
\end{table*}

\begin{table*}[t]
\centering
\small
\caption{Ablations on the preregistered, domain-balanced six-paper subset (two per family). $\Delta$ is relative to an independent re-execution of Full PaperJury on the same subset under matched harness and budget; K follows the Table~\ref{tab:main_results} convention, and w/o bounded review removes the cap.}
\label{tab:ablation_results}
\setlength{\tabcolsep}{5pt}

\begin{tabular}{lcccccccc}
\toprule
Variant & F1 & $\Delta$F1 & Acc v & $\Delta$Acc v & ESVR & $\Delta$ESVR & K & W \\
\midrule
Full PaperJury & 0.649 & --- & 0.881 & --- & 0.029 & --- & 3.00\,$\pm$\,0.63 (0/6) & 2.43 \\
w/o bounded review & 0.572 & $-$0.077 & 0.868 & $-$0.013 & 0.042 & $+$0.013 & 3.67\,$\pm$\,0.52 (n/a) & 4.81 \\
w/o routing & 0.636 & $-$0.013 & 0.806 & $-$0.075 & 0.047 & $+$0.018 & 3.33\,$\pm$\,0.52 (0/6) & 3.49 \\
w/o trial & 0.642 & $-$0.007 & 0.728 & $-$0.153 & 0.052 & $+$0.023 & 3.17\,$\pm$\,0.75 (0/6) & 2.18 \\
w/o claim spine & 0.632 & $-$0.017 & 0.857 & $-$0.024 & 0.112 & $+$0.083 & 3.00\,$\pm$\,0.63 (0/6) & 2.37 \\
w/o guard chain & 0.627 & $-$0.022 & 0.859 & $-$0.022 & 0.181 & $+$0.152 & 2.83\,$\pm$\,0.75 (0/6) & 1.94 \\
\bottomrule
\end{tabular}%

\end{table*}

Table~\ref{tab:ablation_results} isolates the roles of bounded review, deterministic routing, the due-process trial, the claim spine, and the guard chain; its Full PaperJury row is re-executed on the subset, so its aggregates differ slightly from the main run by run stochasticity (e.g., F1 0.649 vs.\ 0.650). The ablations are informative only insofar as degradations align with the function of the removed component, rather than merely showing that the system still produces outputs. The measured pattern matches the component functions: the largest F1 drop comes from removing bounded review ($-$0.077), the largest verdict-agreement drop from removing the due-process trial ($-$0.153), the largest safety degradations from removing the guard chain ($+$0.152 ESVR) and the claim spine ($+$0.083), and removing deterministic routing mainly damages verdict agreement ($-$0.075) while raising cost (2.43 to 3.49 hours per paper).

PaperJury w/o Deterministic Routing primarily tests contestability-based routing. Its measured signature is degraded adjudication allocation together with higher cost: verdict agreement falls by 0.075 and wall-clock time rises from 2.43 to 3.49 hours per paper while F1 barely moves ($-$0.013), because more issues are pushed into semantic adjudication instead of being filtered or handled more lightly. This variant therefore probes whether the critique-to-adjudication boundary depends on deterministic orchestration rather than extra semantic judgment. The same logic clarifies why broader issue generation can increase burden even when it appears to improve coverage: additional weakly grounded critiques may enter the durable ledger and consume expert-like adjudication effort.

PaperJury w/o Due-Process Trial targets verdict and routing quality on contestable substantive-major issues. Replacing whole-paper defense and the decorrelated local-context jury with single-pass semantic judgment should make borderline cases more sensitive to framing and incomplete local evidence. The measured degradation is concentrated exactly there: removing the trial produces the largest verdict-agreement drop ($\Delta$Acc~v $-$0.153) while issue discovery is nearly intact ($\Delta$F1 $-$0.007), consistent with borderline cases needing manuscript-wide context for stable decisions.

The two edit-safety ablations, w/o claim spine and w/o guard chain, directly test edit safety and verified closure. Disabling the frozen claim spine, or the anchor-bounded diff checks, meaning audit, and compile guard chain, lets more patches pass initial screening at sharply degraded safety: ESVR rises to 0.112 and 0.181 against 0.029 for the full system, while F1 and Acc~v move by at most 0.024, and w/o guard chain also terminates fastest (1.94 hours per paper), the signature of unchecked application rather than improved revision. These results support risk-proportional guard chains as a necessary condition for artifact-safe LaTeX revision rather than conservative overhead.

Table~\ref{tab:esvr_companion} reports the mandatory companions to ESVR. PaperJury applies a similar edit volume to the judge-centered loop (13.4 vs.\ 14.3 applied edits per paper) at slightly lower coverage of valid-fixable terminals (0.821 vs.\ 0.843), because the guard chain blocks 17.0\% of proposed patches before application versus 3.4\%; in exchange, the per-edit safety violation rate is 4.4$\times$ lower (4/161 vs.\ 19/172). The forward-only rewriter applies the most edits (21.2 per paper) with by far the worst safety (ESVR 0.240), which is why ESVR is only interpretable jointly with edit volume and coverage.

%% file: sections/conclusion.tex
\section{Conclusion}

This work suggests that pre-submission hardening is better treated as a governed closed-loop process than as open-ended critique generation or forward-only rewriting. In that setting, PaperJury is framed as a deterministic-versus-semantic split in which deterministic orchestration owns state, routing, stopping, and exact-once patch application, while semantic agents remain limited to bounded review, judgment, and repair. Because the current evaluation is designed to test systems along issue quality, verdict and routing quality, edit safety, convergence behavior, and cost, these dimensions remain the main criteria for assessing trustworthy unattended assistance; the explicit invalid-drop, valid-fixable, and author-required outcomes likewise motivate a three-way terminal verdict space when some critiques should be rejected or deferred rather than revised. The approach is still bounded by the expert-review setting and the studied LaTeX computer science manuscript workflow, and related systems have typically emphasized feedback generation, structured review, or broader workflow governance rather than this particular combination of closed-loop adjudication, bounded revision, and deterministic completion. More broadly, durable state, due-process adjudication, and deterministic stopping may provide a useful template for AI-assisted scientific workflows that require accountable end-to-end action.